# A suite of LMs comprehend puzzle statements as well as humans

Adele E Goldberg[1*], Supantho Rakshit[1], Jennifer Hu[2], Kyle Mahowald[3]

[1]Princeton University  [2]Harvard University,  [3]UT Austin

## Abstract

Recent claims suggest that large language models (LMs) underperform humans in comprehending minimally complex English statements (Dentella et al., 2024). Here, we revisit those findings and argue that human performance was overestimated, while LLM abilities were underestimated. Using the same stimuli, we report a preregistered study comparing human responses in two conditions: one allowed rereading (replicating the original study), and one that restricted rereading (a more naturalistic comprehension test). Human accuracy dropped significantly when rereading was restricted (73%), falling below that of Falcon-180B-Chat (76%) and GPT-4 (81%). The newer GPT-o1 model achieves perfect accuracy. Results further show that both humans and models are disproportionately challenged by queries involving potentially reciprocal actions (e.g., kissing), suggesting shared pragmatic sensitivities rather than model-specific deficits. Additional analyses using Llama-2-70B log probabilities, a recoding of open-ended model responses, and grammaticality ratings of other sentences reveal systematic underestimation of model performance. We find that GPT-4o can align with either naive or expert grammaticality judgments, depending on prompt framing. These findings underscore the need for more careful experimental design and coding practices in LLM evaluation, and they challenge the assumption that current models are inherently weaker than humans at language comprehension.

There is much interest in whether Language Models (LMs) are "human-like" on a variety of tasks and capabilities (Mitchell, 2021; Mitchell & Krakauer, 2023; Mahowald, et al. 2024). But it is not actually straightforward how to operationalize and evaluate such claims. Models vary in size, training data, architecture. Experiments can be conducted in English or a host of other languages. Researchers must decide how to code responses and judge accuracy. Comparisons with humans bring another set of design decisions, concerning how data is displayed, whether instructions or examples are provided and how those might influence results, and whether binary, ordinal, or gradient responses are collected  (Hu and Frank, 2024; Lampinen and Dasgupta, 2024; Ivanova, 2025).   In this context, we here revisit a recent report focused on comprehension, Dentella et al. (2024) (DGMML). The report suggests that large LMs struggled to comprehend even minimally complex English statements, in a comparison to humans. Here we argue that certain design and coding decisions overestimated human comprehension and underestimated the performance of LMs. When humans are required to process statements and queries sequentially—without the ability to reread the statement after the query—their performance drops significantly, and they are outperformed by Falcon-180B-Chat (Almazrouei et al. 2023) and GPT-4 (OpenAI, 2023). A newer model, GPT-o1-mini, achieves perfect accuracy. Notably, both humans and lower-performing LMs performed worse on items involving potentially reciprocal actions, suggesting shared sensitivity to pragmatic factors, rather than LM-specific deficits.

The task at issue involves what we refer to as *puzzle stimuli*. Representative instances are provided in Table 1.

**Table 1:** Example puzzle stimuli, including 1-2 statements plus query, and required responses (from DGMML 2024)

| **Puzzle stimuli** (statement and <u>query</u>) | **Target response** |
|---|---|
| (1) Cleo kissed Alice and Alice was kissed by Mary. Cleo and Alice were kissed by Mary. <u>In this context, was Mary kissed?</u> | No |
| (2) Flavia and Jack avoided Mary and Franck was avoided by Lucy and Flavia. <u>In this context, did Franck avoid anyone?</u> | No |
| (3) Alice yelled at Flavia and Flavia yelled at Alice's sister. Alice's sister was greeted by Alice. <u>In this context, was Alice greeted?</u> | No |

Though the stimuli were described as minimally complex by DGMML, we actually find them quite challenging to answer. They include an abundance of proper names, potential confusion from garden path interpretations due to minimal or non-standard punctuation choices, and irrelevant information presented in main clauses as if it were at issue. Particularly odd is the fact that all queries were prefaced by "In this context,' but target responses required that any context-based inferences be strictly ignored. To see this, assume Alice and Cleo were kissed by Mary (as in Table 1 [1]). In this context, did anyone kiss Mary? Optimistic readers will at least hope that Alice or Cleo reciprocated Mary's kiss, yet the target response (No) eclipses that possibility. For instance, the open text response in (1a) to (1) was treated as an error:

(1a)   Yes, Mary was kissed in this context. Firstly, Cleo kissed Alice, indicating that Mary might have witnessed or joined the act. Secondly, Alice was kissed by Mary solidifying the notion that Mary engaged in kissing activity. Lastly, it is mentioned again that Cleo and Alice were kissed by Mary reinforcing the idea that Mary was involved in acts of kissing. Overall, all indications point toward Mary participating in kisses, thus answering the question
affirmatively. --mistralai/Mixtral-8x7B-Instruct-v0.1

A full 40% of stimuli queried actions that are typically reciprocated (kissing, hugging, greeting).
The focus on a particular interpretation of this fairly narrow class of verbs seems unusual for a
broad test of language understanding.

*Human performance drops in more naturalistic design*

For humans to achieve the remarkably high accuracy reported by DGMML (.90), we speculated
that they were allowed to scan back and forth between statement and query. This, however,
undermines the claim that people exhibited strong comprehension of the puzzle stimuli, since
people primarily reread sentences only when they *fail* to adequately comprehend it (Booth &
Weger, 2013; Reyner et al., 2006). To test this possibility, we ran a preregistered study that
collected 1-word (forced choice) responses on the same stimuli from 120 Prolific participants
(mean age = 38.9; 82 female), half of whom read each query immediately after the statement
with no opportunity to reread the statement (sequential condition). The other half read each
statement and query on a single page (simultaneous condition). As predicted, human accuracy
was lower when people are unable to reread the statements (sequential condition: 73.0% vs.
simultaneous condition: 85.2%, $\beta$ = -0.11, $t$ = -5.10, $p <$ .0001) and the sequential condition is
also lower than DGMML's participants' responses ($\beta$ = -0.11, $t$ = -5.01, $p <$ .0001).

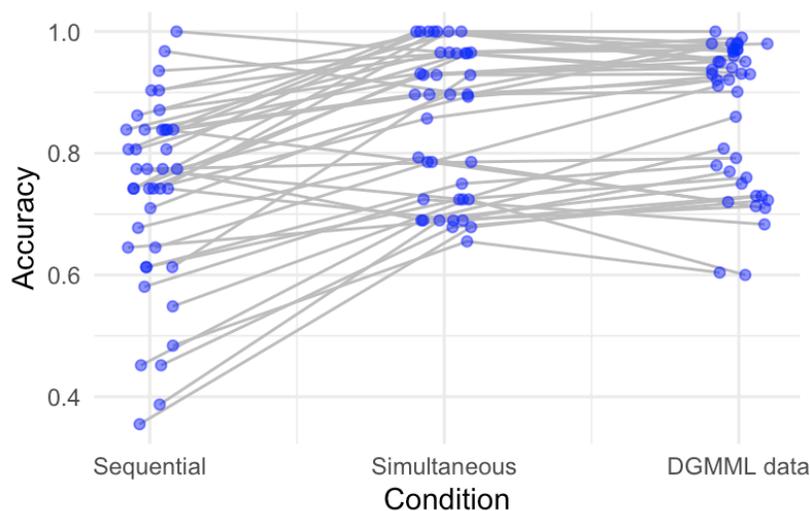

Figure 1: Human accuracy is lower when each query is presented immediately after its premise
statement (sequential condition) than when people are allowed to scan back and forth between
statement and query (simultaneous condition), the latter replicating data collected by DGMML.

Mean accuracy of 1-word responses provided by Falcon-180B-Chat (.76), GPT-4 (.81), and the newer GPT-o1 model (1.0) shows they each outperformed human participants in the sequential condition (Figure 2).

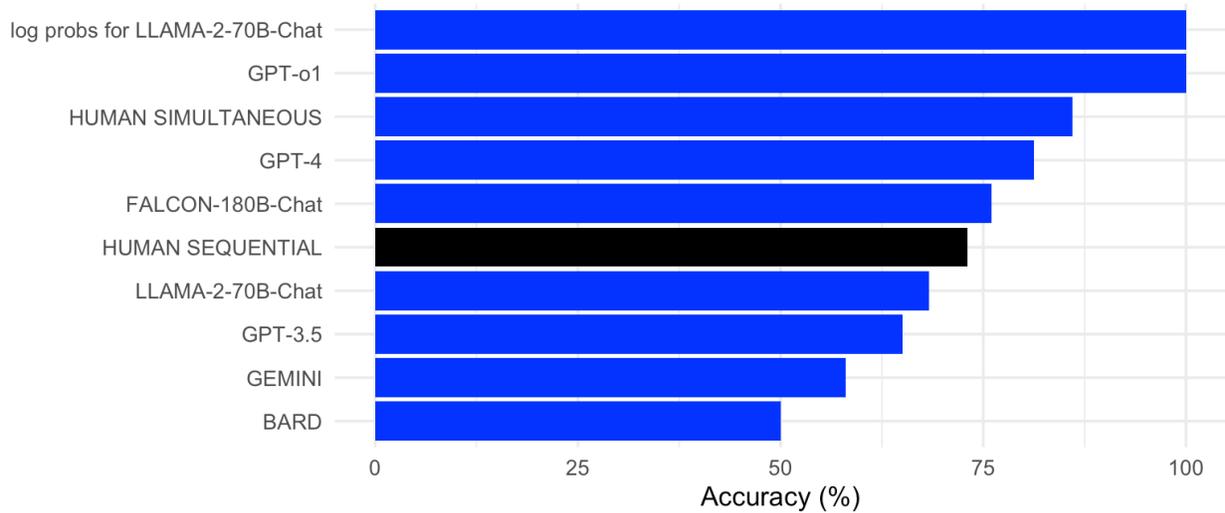

Figure 2: Comparison of model and current human accuracy on target responses of puzzle stimuli . Human performance in the current more naturalistic sequential condition shown in black. Also included are new results from GPT-o1 and an analysis of log probabilities gathered from Llama-2-70B-chat.

Having shown that human performance was overestimated in the original study, we next report three new analyses demonstrating that models align better with human responses than originally indicated based on (1) log probabilities from Llama-2-70b (Touvron et al. 2023), rather than binary responses to explicit probes (2) a comparison of LM and human sensitivity to the same pragmatic factor (3) a recoding of the open-ended LM responses originally obtained by DGMML.[1] Finally, we revisit the general framing of the DGMML report, which implied that models ought to outperform humans in detecting rule-based grammaticality judgments—judgments we demonstrate are not recognized by people.

*Log probabilities of Llama-2-70B reveal full accuracy*
DGMML collected binary judgments, but as Hu and Levy (2023) argue, log probabilities provide a more informative gradient measure of acceptability and avoid relying on models' metalinguistic skill (Hu et al., 2024). Using Llama-2-70B as an illustration, we collected three iterations of log probabilities of the full statements followed by the target responses *vs.* non-target responses. Results show higher log probabilities for target responses on *every item*, indicating far better performance than when binary results of metalinguistic probes were used.

---

[1] We do not collect reliability estimates since these can be expected to vary for LMs on the basis of temperature setting and whether 0-shot learning is used. Reliability will also vary for people depending on motivation, fatigue, whether repetitions are interpreted as opportunities to correct errors, and time delays, as well as the actual stability of comprehension. LM responses collected by DGMML indicate that 0-shot learning was not used in probes.

*Models and humans are sensitive to the same pragmatic implication*

As expected, participants produced more non-target responses ('errors') to stimuli containing potentially reciprocal actions (kissing, hugging, greeting) in both the newly collected human results. This was confirmed by generalized linear models on each human data set, with subjects and items as random intercepts: (sequential condition: $\beta = -1.0$, $t = -3.87$, $p < .0001$; simultaneous condition $\beta = -1.67$, $t = -4.09$, $p < .0001$) and in DGMML's human dataset ($\beta = -1.7$, $t = -3.75$, $p < .0001$).

Turning our attention to LM performance, we note that it is nonsensical to combine data across different models since models vary in size, training data, and architecture. Generalized linear models that include reciprocal verbs as a fixed factor reveal that target responses by the lowest performing models–Bard and Gemini–are challenged by items that included potentially reciprocal actions, as is the case for humans (Gemini: $\beta = -2.01$, $z = -2.93$ $p = 0.003$; Bard $\beta = -2.16$; $z = -3.118$ $p = 0.002$). Using the more informative likelihood estimates of target vs non-target responses as the outcome for Llama-2-70b reveals that it, too, is less certain of its responses when queries involve potentially reciprocal actions ($\beta = -1.38$, $t = -3.23$ $p = 0.002$).

*Revisiting the coding of open-ended responses*

DGMML report that accuracy for open-ended responses was lower than that of binary responses. Here we note that coding of accuracy is subjective, as already indicated in the introduction (recall example [1a]). A re-coding of LM's open-setting responses by two coders before adjudication reveals that 25% of responses were scored either ungenerously or incorrectly in that: a non-target response was provided with sound justification (e.g., recall [1a]); the target response was provided, but with certain extraneous inaccurate material; the response was wholly appropriate but scored as an error for unclear reasons.[2]

*Human and model grammaticality judgments vs. idealized grammaticality judgments*

Before concluding, we observe that DGMML framed their empirical work by observing that two models, GPT-3 and GPT-3.5, failed to provide the intended binary grammaticality judgments on a handful of sentences. We queried our human participants on these sentences, and

---

[2] An example coding error is the following response to the first stimulus in Table 1 ([1]) provided by llama-2-70B:

(i) No. The sentences "Cleo kissed Alice" and "Alice was kissed by Mary. Indicate that Mary was the recipient of the kisses, not the one performing the kisses. Therefore, Mary was not kissed It's important to note that the phrase "kissed by" implies that the person mentioned before the verb ("by") is the one performing the action, while the person mentioned after the verb ("Mary") is the recipient of the action. So in this case, Mary is the recipient of the kisses, and therefore she was not kissed.

additionally probed GPT-4o (OpenAI, 2024) on new versions of the stimuli in two ways, as reported below.

DGMML's article opens with a table of example sentences, here provided in Table 2 (right column), that are unrelated to the puzzle stimuli. No human judgments on these sentences were reported, although target responses as indicated in the far-right column of Table 2, were stipulated. Green color is used to signify grammaticality and red, ungrammaticality. DGMML reported that GPT-3 and GPT-3.5 failed to match the target responses on the majority of sentences (column 2, Table 2).

To empirically verify target responses, we asked the same participants who took part in either puzzle stimuli survey to judge each sentence on a Likert scale from 1 = fully grammatical to 7 = fully grammatical. Mean responses are provided in the Table 2 (column 3), with color assigned according to whether the mean fell above (green) or below (red) the middle of the scale. Human judgments aligned with GPT-3 and GPT-3.5's classifications on the majority of sentences (5/7) and diverged from the proposed target responses in 6 out of 7 sentences. The only case in which humans aligned better with the target responses than the models, was the last item (g): people recognize that "The doctor the nurse the clinic had hired met Jack" is ungrammatical.

We also probed GPT-4o, using a Likert scale with 1= totally ungrammatical and 7 = totally acceptable, on newly created sentences that shared the same features as the items in 1-7, to mitigate contamination concerns (see SI). Only one item (d) was reversed in its judgment from the earlier GPT models: i.e., GPT-4o detected a lack of agreement between subject and verb that earlier models missed. In a final probe of the same newly created items, we specified that we wanted the ratings that "professional linguists would provide." With these instructions, GPT-4o's responses aligned with the target responses proposed by DGMML on 5/7 items.

Table 2. Responses, coded by color as grammatical (green) or ungrammatical (red) from probes of GPT-3, GPT-3.5 along with new empirical human judgments and DGMML's target responses (2024: Table 1). Also new are judgments on newly created sentences (see SI), provided by GPT-4o under two sets of instructions: a) the same instructions provided to people and b) judgments when asked to respond as a professional linguist would.

| Sentence Stimuli | GPT-3 and GPT-3.5 | New Human ratings $N$ = 120 | New: GPT4-o ratings on revised items | New: GPT4-o ratings "as a professional linguist" | DGMML's proposed target responses |
|---|---|---|---|---|---|
| a. Dogs dogs dog dog dogs. | | 2.08 | 2 | 7 | 'grammatical' |
| b. Fish fishermen catch eat worms. | | 2.19 | 3 | 7 | 'grammatical' |
| c. The key to the drawers <u>are</u> on the table. | | 4.71 | 2 | 2 | 'ungrammatical' |
| d. The village of the fundamentally flawed ideas have not met the criteria. | | 3.76 | 2 | 2 | 'ungrammatical' |
| e. More people have been to Russia than I have. | | 4.85 | 5 | 6 | 'ungrammatical' |
| f. The patient the nurse the clinic had hired admitted met Jack. | | 1.44 | 3 | 7 | 'grammatical' |
| g. The doctor the nurse the clinic had hired met Jack. | | 1.71 | 5 | 7 | 'ungrammatical' |

*Discussion*

To summarize, the design and coding choices made by DGMML arguably overestimated human responses and underestimated LM responses. We have shown that when humans are faced with a more natural comprehension task, they actually display performance within the middle range of models' varied performance.

**Conclusion**

As LM models advance, it becomes harder to identify language tasks that no model can perform well on, or, at least as well as an average human. Vigilance is required to avoid underestimating the potential of LMs or overestimating our own uniqueness. Here we replicate DGMML's finding of strong performance by humans on their adversarial comprehension task, but only when people are allowed to scan back and forth between the puzzle stimuli and the query. When we mitigate the potential for rereading by presenting each query immediately after the stimulus, human accuracy drops below that of Falcon-180B-chat and GPT-4. The newer GP-o1 model shows perfect accuracy, as does Llama-2-70b, when log probabilities of target and non-target responses are compared. Moreover, Llama-2-70b's log probabilities reveal lower confidence on items that involve reciprocal actions, a vulnerability also evident in human errors (and those made by Bard and Gemini).

Finally, we demonstrate that human grammatical judgments differ from idealized judgments suggested by DGMML. Instead, human judgments align better with GPT-3's and GPT-3.5's responses. Notably, GPT-4o was able to provide judgments that align well with naive human judgments *and* it adjusted its judgments to align closer to the idealized judgments when asked to provide "judgments that professional linguists would give."

A question worth raising in this context is, under what conditions ought we to expect LMs to provide idealized responses rather than human-like responses? There may be certain situations where it is important to respond to puzzles in a categorical manner that ignores contextual inferences. Current results show that certain models already do this *better* than the average person.

**Supplementary Information**

*Stability of responses across repetitions*

We did not include repetitions of items in our surveys because stability of responses in humans are influenced by a variety of factors, including: motivation, fatigue, and whether repeated items are interpreted to result from error. For LMs, when 0-shot learning is used, stability is determined by the temperature setting. When 0-shot learning is not used, LM stability, like human stability, may depend on whether it interprets repetitions as an indication that the previous response was inadequate.

**Table SI.1**: Edited stimuli used to probe GPT-4o for grammaticality judgments and full

| | |
|---|---|
| *a. Rats rats rat rat rats.* | 7/7 Fully grammatical under recursive nominal and verbal structures (e.g., "rats [that] rats [who] rat [other] rats rat, rat rats"). Processing difficulty is high, but no violation of syntax. |
| *b. Salmon salmon fishers catch eat minnows.* | 7/7 Fully grammatical under recursive nominal and verbal structures |
| *c. The price of the tables are on the table.* | 2/7 Ungrammatical due to subject-verb agreement: "The price ... is on the table." |
| *d. The town of the extremely reclusive leaders have not gathered recently.* | 2/7 Same issue: "town" is singular, so "has not gathered" is required. Ungrammatical |
| *e. More dogs have been to the park than Charlie has.* | 6/7 Structurally ambiguous, but grammatically fine. Ambiguity doesn't entail ungrammaticality. |
| *f. The student the teacher the school had hired accepted saw Kelly.* | 7/7 Fully grammatical, though with deep center-embedding. Difficult to parse, but no syntactic error |
| *g. The student the teacher the school had hired saw Kelly.* | 7/7 Same as above: grammatical, just complex. Shorter embedding makes it slightly more acceptable than #6. |